\documentclass[preprint,12pt]{elsarticle}

\usepackage{amssymb}
\usepackage{multirow}
\usepackage{booktabs}
\usepackage{float}
\usepackage{graphicx}
\usepackage[normalem]{ulem}
\usepackage{makecell}
\useunder{\uline}{\ul}{}
\usepackage{array}
\usepackage{bbding}
\usepackage{utfsym}
\usepackage{pifont}
\usepackage{indentfirst}
\usepackage{hyperref}

\begin{document}
\begin{frontmatter}



\title{Dynamic Scene Deblurring Based on Continuous Cross-Layer Attention Transmission}

\author{Xia Hua%
\fnref{1}}

\author{Mingxin Li%
\fnref{1}}

\address[1]{{School of Electrical and Information Engineering},
{Wuhan Institute of Technology},
{Wuhan 430205},
{China}}

\address[2]{{School of Automation},
{Huazhong University of Science and Technology},
{Wuhan 430074},
{China}}


\cortext[cor1]{Corresponding author at: School of Electrical and Information Engineering, Wuhan Institute of Technology,Wuhan 430205,China}

\author{Junxiong Fei\corref{cor1}%
\fnref{1}}
\ead{junxiongfei@gmail.com}

\author{Jianguo Liu
\fnref{2}}

\author{Yu Shi%
\fnref{1}}

\author{Hanyu Hong\corref{cor1}%
\fnref{1}}
\ead{hhyhong@wit.edu.com}


\begin{abstract}
 The deep convolutional neural networks (CNNs) using attention mechanism have achieved great success for dynamic scene deblurring. In most of these networks, only the features refined by the attention maps can be passed to the next layer and the attention maps of different layers are separated from each other, which does not make full use of the attention information from different layers in the CNN. To address this problem, we introduce a new continuous cross-layer attention transmission (CCLAT) mechanism that can exploit hierarchical attention information from all the convolutional layers. Based on the CCLAT mechanism, we use a very simple attention module to construct a novel residual dense attention fusion block (RDAFB). In RDAFB, the attention maps inferred from the outputs of the preceding RDAFB and each layer are directly connected to the subsequent ones, leading to a CCLAT mechanism. Taking RDAFB as the building block, we design an effective architecture for dynamic scene deblurring named RDAFNet. The experiments on benchmark datasets show that the proposed model outperforms the state-of-the-art deblurring approaches, and demonstrate the effectiveness of CCLAT mechanism. The source code is available on: https://github.com/xjmz6/RDAFNet.
\end{abstract}

\end{frontmatter}


\section{Introduction}

Motion blur, caused by camera shaking or a rapid motion of objects in a dynamic scene, is one of the most common artefact  types created when taking photo\cite{Fergus2006}. Image deblurring, that is, restoring a clear image from a blurred image, has always been an important task in image processing and computer vision\cite{Tao2018}. Since the latent sharp image and blur kernel cannot be determined from the observed blurry image, single image deblurring is an ill-posed problem.

To solve this problem, most traditional methods attempt to model, a priori to characterize the features of clear images\cite{peng2020joint,li2019new,shao2016regularized}, such as the prior for natural images\cite{Michaeli2014}, the prior for text images\cite{Pan2014}, and the prior independent of image types\cite{Pan2017}, etc. However, the statistical prior modelling method has a limited ability to describe the features of sharp images, which leads to the corresponding deblurring algorithms being unable to obtain a good clear image. In addition, optimization problems need to be solved after modelling the prior. If the prior is designed to be very complex, it often leads to difficulty in solving the deblurring model \cite{Gong2017}.

Recently, several CNN-based deep learning methods have started to learn end-to-end mapping for deblurring\cite{Li2020,Cai2020,Jung2021,Li2022,Nah2017}. This kind of method does not need to estimate the blur kernel and reduces the error caused by an inaccurate kernel estimation. Since Nah \cite{Nah2017} designed a residual block (Resblock) as the building block of their model, the Resblock (Fig. 1(a)) has become the fundamental block of many CNN-based deblurring networks\cite{Gao2019,Park2020,Zhang2019}. In deep CNN, the hierarchical features which extracted by the convolutional layers in different depths are able to capture different characteristics\cite{Tang2019,Zhang2020}, and the effective use of hierarchical features can provide more clues for image restoration\cite{Wang2018}. The residual dense block (Fig. 1(b)) based on dense connected hierarchical features, has been proposed for high-quality image restoration\cite{Wang2018}. However, the above CNN-based deblurring methods use the same weights for different spatial positions of blurred images, and lack adaptive blurring-feature sensing mechanisms to handle the non-uniform blurred in real dynamic scenes.

With the development of deep learning technology in the field of computer vision, researchers began to design network structures for non-uniform blurred image restoration. The visual attention mechanism can locate the target region in the image and capture the features of the region of interest, which has been successfully applied to detection\cite{Woo2018}, \cite{Hu2018} and classification problems\cite{Hu2018a}, \cite{Purohit2020}. Since the blur of a non-uniformly blurred image varies in different spatial locations, and attention can adaptively learn the similar blur characteristics of different positions, adopting attention mechanism can effectively improve the non-uniform deblurring performance of CNN-based models\cite{Chan1998}. Although the CNNs with attention mechanism have shown their superiority in removing non-uniform blur in dynamic scenes, in most of these networks, the attention maps of different layers are separated from each other, which does not make full use of the attention information from different layers in CNN.

\begin{figure}[t]
	\centering
	\begin{minipage}{0.38\linewidth}
		\centering
		\includegraphics[width=1.\linewidth]{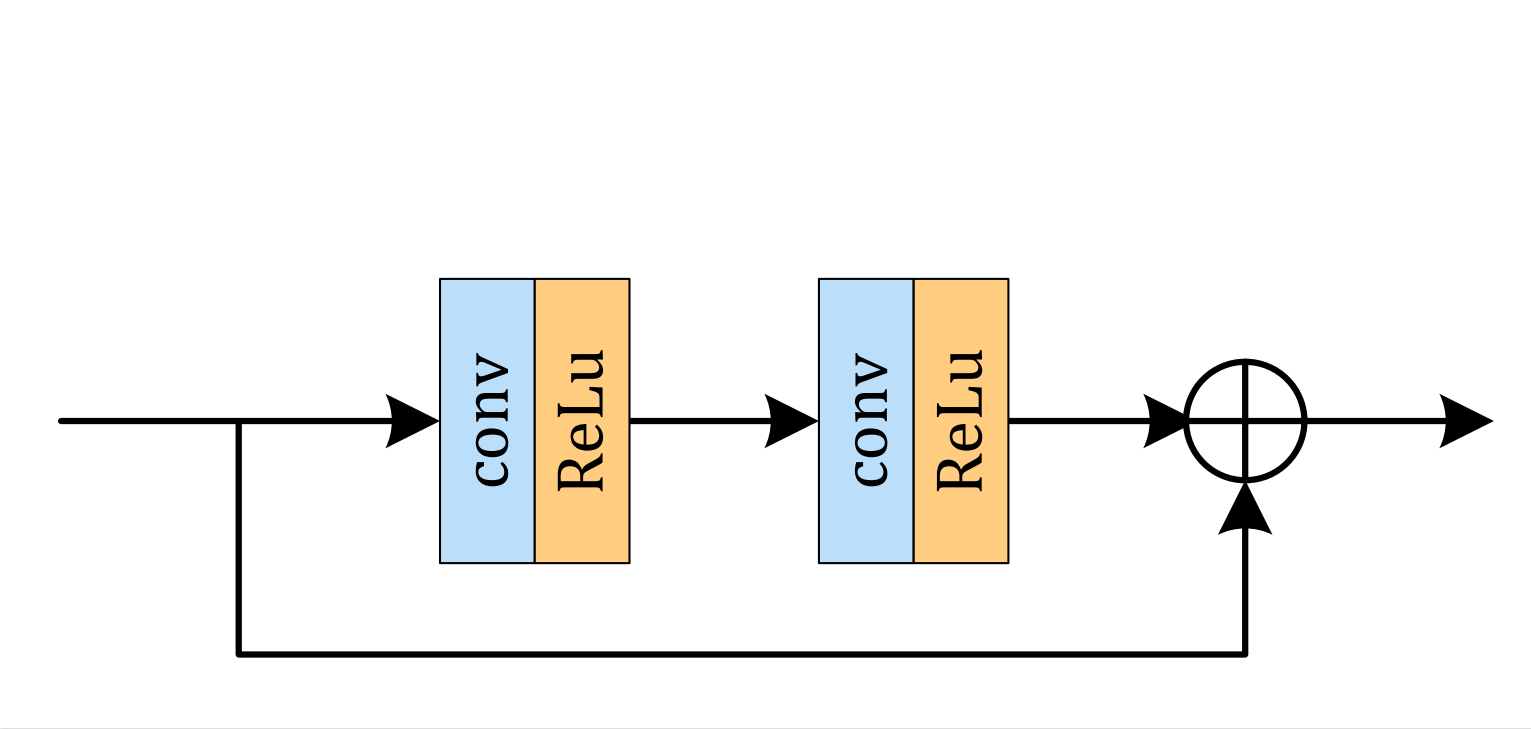}
  \centerline{\footnotesize (a)  ResBlock}
		\label{chutian1}
	\end{minipage}
	\begin{minipage}{0.6\linewidth}
		\centering 
		\includegraphics[width=1.\linewidth]{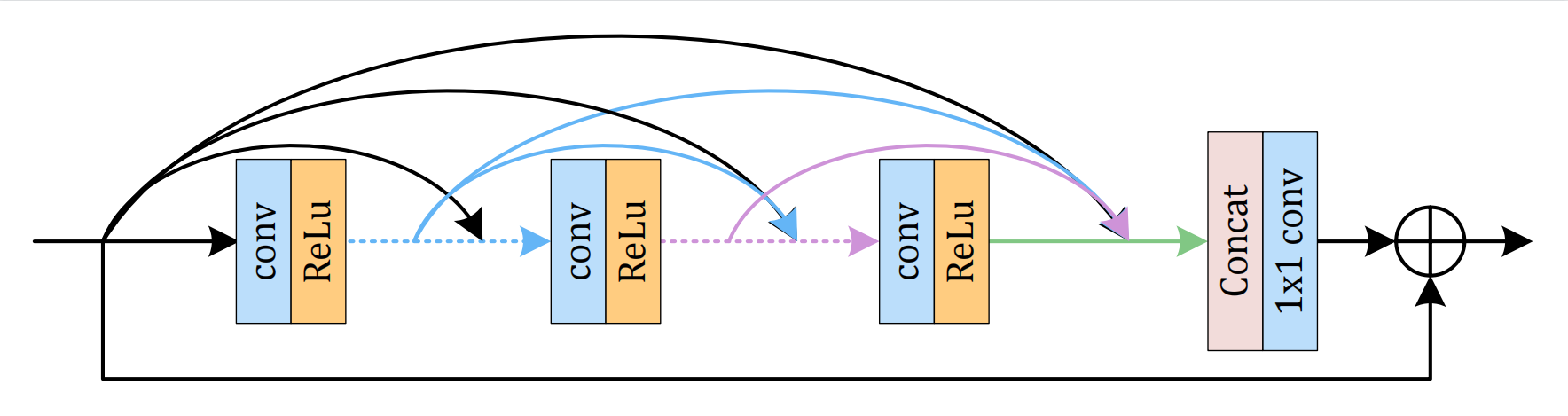}
    \centerline{\footnotesize(b) Residual Dense Block}
		\label{chutian2}
	\end{minipage}
 \begin{minipage}{\linewidth}
		\centering
		\includegraphics[width=1\linewidth]{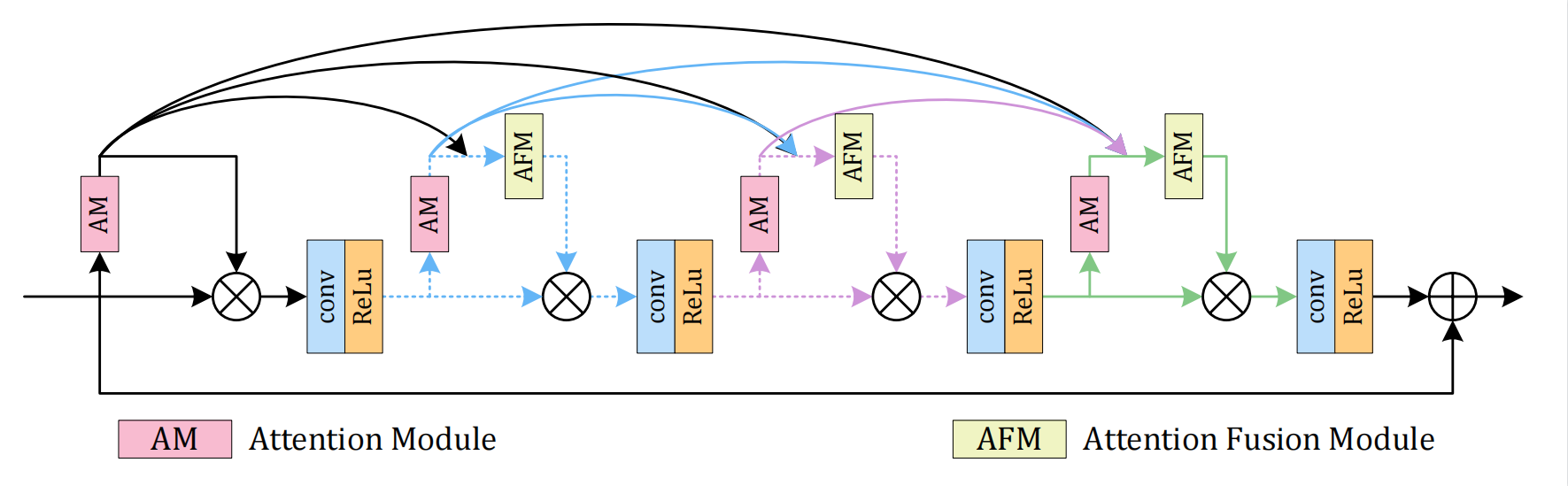}
    \centerline{\footnotesize(c) Residual Dense Attention Fusion Block}
		\label{chutian3}
	\end{minipage}
 \caption{\footnotesize Comparison of prior blocks and our proposed block. The fundamental block in our networks is the RDAFB, which is designed based on the continuous cross-layer attention transmission (CCLAT) mechanism.}
 \label{fig:1}
\end{figure}

Since the hierarchical features would contribute to better image restoration \cite{Zhang2020}, the attention process could focus on the important parts of the features and the attention map contains some highly task-relevant information \cite{Wang2018}, we introduce a new continuous cross-layer attention transmission (CCLAT) mechanism that can exploit the attention information from all the convolutional layers to make better use of hierarchical attentions (We denote the attention maps which are inferred from hierarchical features as hierarchical attentions.). Based on the CCLAT mechanism, we propose a novel residual dense attention fusion block (RDAFB) (Fig. 1(c)). In RDAFB, the attention maps of preceding layers are densely connected with the current attention map, and then the connected attention maps are combined by the convolutional layers to obtain the refined current layer attention map. Taking RDAFB as the building block, we design an effective architecture named RDAFNet.

The main research contributions of this paper are as follows:

\begin{enumerate}[\textbullet]
\item We introduce a new continuous cross-layer attention transmission (CCLAT) mechanism in CNN, which makes full use of all hierarchical attentions through locally dense connections of attention maps.

\item Based on CCLAT mechanism, we just use a simple attention module to construct a novel residual dense attention fusion block (RDAFB), and employ it as a building block to design an effective architecture named RDAFNet for dynamic scene deblurring.

\item Our experiments on benchmark datasets show that the proposed model outperforms the state-of-the-art methods and demonstrate the effectiveness of CCLAT mechanism. 

\end{enumerate}

\section{Related work}

In this section, we briefly introduce relevant work, including statistical prior based methods, deep learning-based method, and attention mechanism.

\subsection{Conventional Methods}

The blind image deblurring is an ill-posed problem, and most traditional methods solve this problem by imposing constraints on blur kernels or latent images\cite{Pan2014}, \cite{Pan2017}. Thus, a large number of effective priors have been proposed, such as sparse gradients distribution model\cite{Chan1998,Levin2009,Shan2008,Xu2013}, hyper-Laplacian prior \cite{Krishnan2009}, dark channel prior\cite{Pan2017}, extreme channel prior\cite{Yan2017}, Local Maxi-mum Gradient (LMG) prior\cite{Chen2019}, etc. Most of the blind image deblurring methods focus on solving the blur caused by camera movement, while the real dynamic scene includes camera movement, rigid or non-rigid object movement, and the variation of scene depth. These methods are difficult to deal with the blur in dynamic scene.

\subsection{Deep Image Deblurring}

In recent years, deep learning methods have been applied to image and video deblurring problems. Through a large amount of training data and learnable parameters, the features of clear images are described, and clear images are directly estimated in an end-to-end way \cite{Tao2018},\cite{Park2020}, \cite{Zhang2019}, \cite{chen2021hinet,Cho2021,Zou2021}. Some researchers have designed multistage-based networks for image deblurring \cite{Tao2018}, \cite{Nah2017}, \cite{Zhang2019}, \cite{Suin2020}. Nah et al. \cite{Nah2017}, inspired by traditional coarse-to-fine approaches, proposed a multiscale deep convolutional neural network and adopted the multiscale loss function to constrain the network training process. This method can effectively solve the problem of image deblurring in dynamic scenes to some extent. Inspired by the above method, Tao et al. \cite{Tao2018} proposed the scale-recurrent network structure to solve the problem of image deblurring in dynamic scenes. The network structure is also a multiscale architecture that uses a ConvLSTM \cite{Hochreiter1997}, \cite{Shi2015} module at different scales to fully extract useful features to remove blur. In the multiscale models, subsampling the original image into a low-resolution image often leads to the loss of spatial information, which is not conducive to image restoration\cite{Brehm2020}. To solve these problems, Zhang et al. \cite{Zhang2019} proposed a simple and effective deep hierarchical multipatch network, which split the blurred image into multiple nonoverlapping patches as input. These networks use the same weight in different spatial positions, while non-uniform blur is spatially-varying, which cannot satisfy the removal of non-uniform blur in real dynamic scenes.

\begin{figure}[t]
  \centering
   \includegraphics[width=0.98\linewidth]{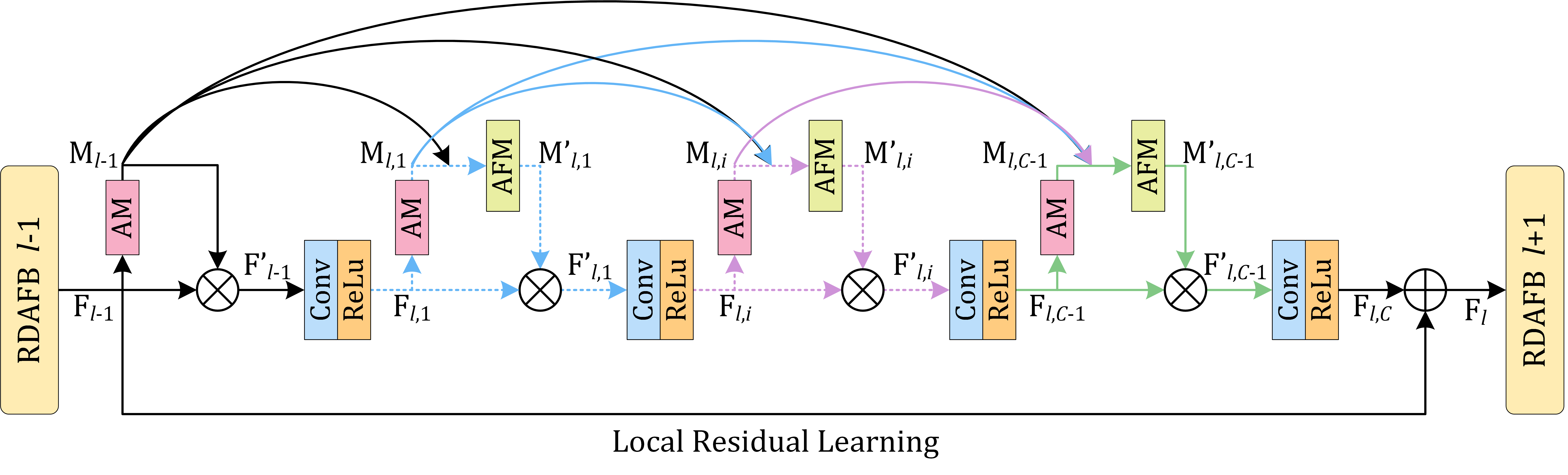}
    \caption{\footnotesize Proposed Residual Dense Attention Fusion Block (RDAFB) architecture. According continuous cross-layer attentional transmission mechanism, the attention maps of preceding layers are densely connected with the current attention map, and the connected attention maps are fused to obtain the refined current attention map.}
      \label{fig:2}
\end{figure}

\subsection{Attention Mechanism}

Attention in human perception usually refers to the ability of the human visual system to process visual information adaptively and focus on interest regions\cite{Corbetta2002}, \cite{Itti1998}. Many deep learning methods improve the performance by exploiting the effect of attention in networks \cite{Wang2018}, \cite{Woo2018}, \cite{Zhang2018}, \cite{Zhao2020}. Hu et al. \cite{Hu2018a} proposed the Squeeze-and-Excitation (SE) block, which can model the interdependence between the channels, adaptively recalibrates the feature response of each channel and achieves remarkable performances in image classification tasks. Zhang et al. \cite{Zhang2018} introduced the residual channel attention block (CAB) and designed deep residual channel attention networks (RCAN) to achieve better SR performance than the traditional CNN-based methods. Zamir et al. use CAB as fundamental block to design a multistage architecture named MPRNet, which has achieved significant restoration results \cite{Zamir2021}. 

Some researchers have explored both spatial and channel dimension feature correlations to obtain better performance. Woo et al. \cite{Woo2018} proposed a simple and effective convolution block attention module (CBAM) in which the attention maps can be derived along the spatial and channel dimensions, and then the attention maps are multiplied by the original feature to refine the features. Dai et al. \cite{Dai2019} proposed a residual block attention network (RBAN) framework, which effectively utilizes feature correlation in the spatial and channel dimensions to enhance feature expression. Recently, self-attention (SA) \cite{Vaswani2017} has been widely used to advance the field of image processing \cite{Zhang2019a} and computer vision \cite{Wang2018} because the SA module can explore long-range dependencies at the pixel level. Some works \cite{Chen2021,Zamir2022,Wang2022} began to use a self-attention layer in the low- resolution features to achieve better restoration results, but SA requires a lot of memory \cite{Zamir2022}. 

The above methods show that the attention mechanism has some advantages in the removal of non-uniform blur in dynamic scenes. However, some attention modules in these networks are very complex, and each layer of attention map is used only to refine the current features. To address this problem, we introduce a new continuous cross-layer attention transmission (CCLAT) mechanism to make full use of the attention information from different layers in CNN for dynamic scene deblurring.

\section{Proposed Approach}

In this section, the details of CCLAT mechanism, RDAFB and RDAFNet are provided in the following subsections.

\subsection{Continuous Cross-Layer Attention Transmission Mechanism.} 

Our goal is to fuse the attention information from different layers as much as possible, but directly fusing the attention maps from all layers would stack huge amount of attention information. Instead, we introduce a new continuous cross-layer attention transmission mechanism, which first adaptively fuse the attention information locally and then transmit it to the subsequent attention fusions. It is implemented by passing the attention map inferred from the output features of preceding RDAFB to each layer of the current RDAFB. Let $F_{l-1}$ and $F_{l}$ be the input and output features of the l-th RDAFB respectively. To suppress those features with less information at preceding (l-1)-th RDAFB and allow the useful ones to be transmitted to the l-th RDAFB, we firstly generate attention map  and get the refined feature 

\begin{figure}[!h]
	\centering
	\begin{minipage}{\linewidth}
		\centering
		\includegraphics[width=1.\linewidth]{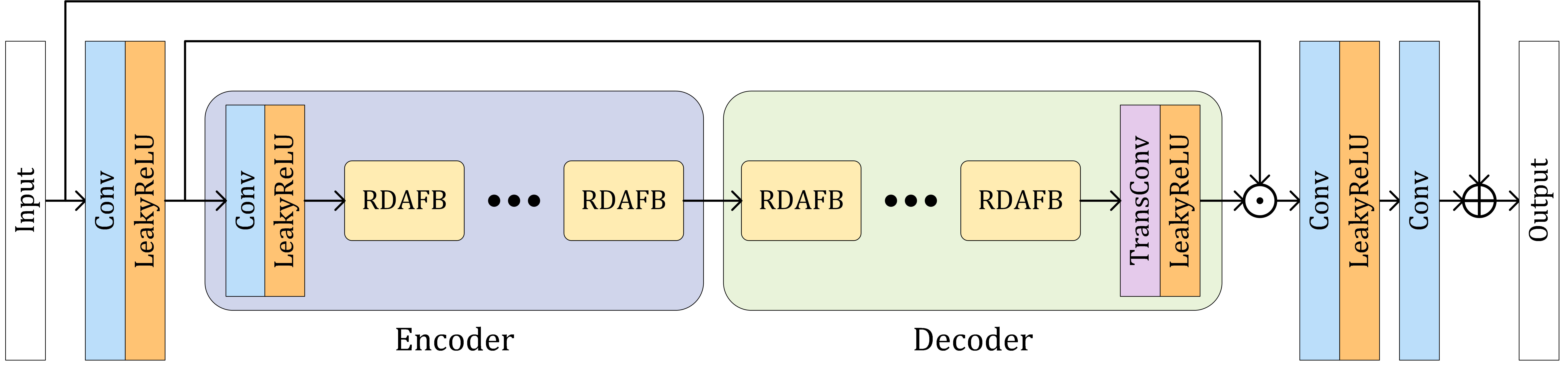}
  \centerline{\footnotesize (a) Single-stage RDAFNet}
		\label{chutian1}
	\end{minipage}
	\begin{minipage}{\linewidth}
		\centering 
		\includegraphics[width=1.\linewidth]{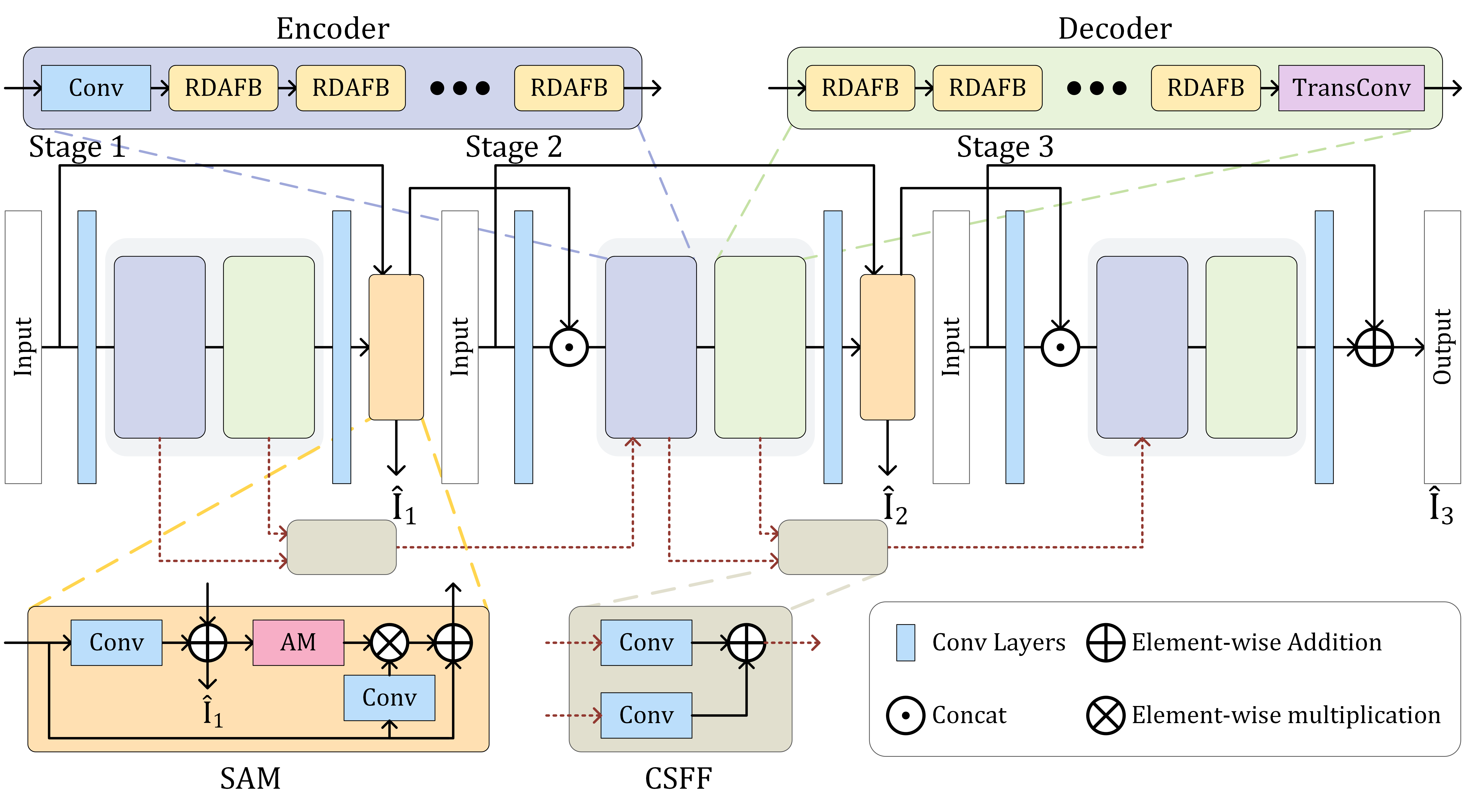}
    \centerline{\footnotesize(b) Multi-stage RDAFNet}
		\label{chutian2}
	\end{minipage}
\caption{\footnotesize The Architecture of proposed Residual Dense Attention Fusion Network (RDAFNet). The single-stage RDAFNet uses one downsampling and one upsampling operations, and the encoder-decoder of single-stage RDAFNet is stacked by RDAFBs. The multi-stage RDAFNet consists of three sub-networks, each subnetwork is similar to that of single-stage RDAFNet. }
\label{fig:3}
\end{figure}

\begin{equation}
    M_{l-1}=f_l\left (F_{l-1}\right ),F^{'}_{l-1}=M_{l-1} \otimes F_{l-1}
\end{equation}
where $f_l$ is the function of first attention module in l-th RDAFB, and $\otimes$ denotes elementwise multiplication. To transmit the attention information of the previous layer to the current layer, we design an attention fusion module. Using this module, the attention maps of preceding layers are densely connected with the current attention map, and the connected attention maps are fused to obtain the refined current attention map. The output of i-th attention fusion module can be obtained by
\begin{equation}
    M^{'}_{l,i}=g_{l,i}\left (M_{l-1},M_{l,1},\cdot\cdot\cdot,M_{l,i}\right )
\end{equation}
where $g_{l,i}$ is the function of i-th attention fusion module in l-th RDAFB.$M_{l,i}$ is the $\left(i+1\right)$-th attention map which can be obtained by
\begin{equation}
    M_{l,i}=f_{l,i}\left (F_{l,i}\right )
\end{equation}

where $F_{l,i}$ and $f_{l,i}$ denote the feature extracted from the i-th convolution layer and the function of $\left(i+1\right)$-th attention module in the l-th RDAFB respectively. After we get the i-th fused attention map $M^{'}_{l,i}$, the feature $F_{l,i}$ can be refined by
\begin{equation}
    F^{'}_{l,i}=M^{'}_{l,i} \otimes F_{l-1}
\end{equation}
With CCLAT mechanism, the attention maps inferred from the outputs of the preceding RDAFB and each layer are directly connected to the subsequent ones, which not only preserves the feed-forward nature of the network but also extracts local dense attention information.
%
\subsection{Residual Dense Attention Fusion Block}

According to the CCLAT mechanism, we use a simple attention module to construct a residual dense attention fusion block (RDAFB). As shown in Figure \ref{fig:2}, our RDAFB contains several attention modules (AMs), attention fuse module (AFMs) and local residual learning (LRL), leading to a continuous cross-layer attention transmission (CCLAT) mechanism. In the proposed RDAFB, all the intermediate attention maps are generated from hierarchical features by several AMs, each intermediate attention map is directly connected to the subsequent ones, and each connected intermediate attention maps are fused by an AFM to obtain the refined current layer attention map which used to recalibrate the current layer features. The details of attention module, attention fusion Module and local residual learning are described as follows.

1) Attention Module: The visual attention mechanism can locate objects in images and capture the features of areas of interest. The goal of deblurring is to restore the blurry part of the image and make it sharp. Therefore, it is extremely important to introduce an attention module in the network that can capture the information of the blurred region and its surrounding structure for the removal of non-uniform blur. Because extensive use of the attention module in networks may significantly increase the overhead of memory and computation, we use a lightweight pixel attention module from \cite{Zhao2020} to design our building block. For the l-th RDAFB, according to this attention module, the attention map $M_{l,i}$ is generated from the intermediate feature $F_{l,i}$ by applying a convolution layer and sigmoid activation function.  This attention module is very simple, and its overhead is low. In other words, this lightweight pixel attention is computed as
\begin{equation}
    AM\left(F_{l,i}\right)=\sigma \left(f_{1\times1} \left(F_{l,i}\right) \right)
\end{equation}
where $f_{1\times1}$ denotes a convolution operation with the kernel size of $1\times1$, and $\sigma$ is the sigmoid activation function.

2) Attention Fusion Module: To complete the CCLAT mechanism, we also design an attention fusion module to fuse the hierarchical attention maps. For the l-th RDAFB, given the hierarchical attention maps $M_{l,1},\cdot\cdot\cdot,M_{l,i-1}$ and $M_{l,i}$, the fused attention map $M_{l,i}$ can be formulated as follows:
\begin{equation}
    AFM\left(M_{l,1},\cdot\cdot\cdot,M_{l,i-1},M_{l,i}\right)=D_{3 \times 3}\left( g_{1 \times 1} \left(Cat\left( M_{l,1},\cdot\cdot\cdot,M_{l,i-1},M_{l,i}\right )\right)\right)
\end{equation}
where $Cat\left(\cdot \right)$ is used to concatenate the attention maps in the channel dimension,$g_{1 \times 1}\left(\cdot \right)$denotes a convolution operation with the kernel size of $1 \times 1$, and $D_{3 \times 3}\left(\cdot \right)$ represents a depthwise separable convolution operation \cite{Chollet2017} with the kernel size of $3 \times 3$, which further reduces the complexity. 

3) Local Residual Learning:  Based on previous\cite{Zhang2020} studies, we also introduce local residual learning into RDAFB to further improve the information flow. The final output of the l-th RDAFB can be expressed as:
\begin{equation}
    F_{l}=F_{l-1}+F_{l,C}
\end{equation}

where C denotes the number of convolution layers in RDAFB, and $F_{l,C}$ represents the feature transformed by the last convolution layer (C-th convolution layer) in the l-th RDAFB.

\subsection{Network Structure}

Using RDAFB as the building block, we further design an effective framework called RDAFNet, which has two variants: the single-stage RDAFNet and the multi-stage RDAFNet, as shown in Figure \ref{fig:3}. Since the subnetwork structure of multi-stage RDAFNet is similar to that of single-stage RDAFNet, we only introduce the multi-stage RDAFNet structure in detail below.

The multi-stage RDAFNet consists of three subnetworks. Each stage of RDAFNet begins with two convolutional layers, where the first layer with a kernel size of $7 \times 7$ is used for extracting the initial features and the second layer employs a stride of two for downsampling. Next, we stack several RDAFBs to fuse the attentions from the preceding convolution layers. At the end of each subnetwork, we first use one transposed convolution to upsample the resulting features to the original size and then use a $7 \times 7$ convolution to obtain the residual output of the deblurring image.

In our multi-stage RDAFNet, the supervised attention module (SAM) and cross-stage feature fusion (CSFF) which come from \cite{Zamir2021} are employed to connect the three subnetworks. By introducing the SAM and CSFF modules, the features of the previous stage can be more effectively transferred to the next stage, relieving information loss. Adopting a multistage architecture is more efficient than a single-stage architecture.Considering that splitting the image into nonoverlapping patches will lead to the discontinuity of the image’s texture information \cite{chen2021hinet}, we use the original-resolution image as the input of each subnetwork.

\begin{figure}[!h]
\centering
	\begin{minipage}{0.24\linewidth}
		\centering
		\includegraphics[width=1\linewidth]{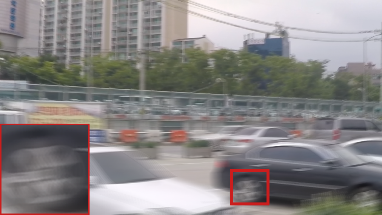}
            \centerline{\footnotesize (a) Blurred image}
	\end{minipage}
	\begin{minipage}{0.24\linewidth}
		\centering
		\includegraphics[width=1\linewidth]{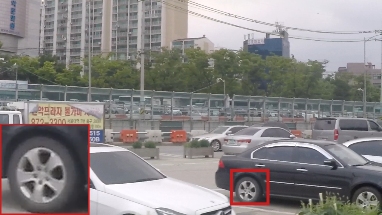}
            \centerline{\footnotesize (b) Sharp}
	\end{minipage}
        \begin{minipage}{0.24\linewidth}
		\centering
		\includegraphics[width=1\linewidth]{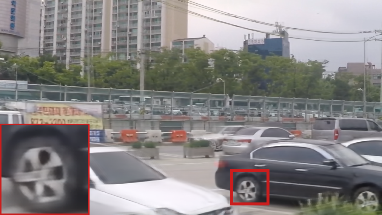}
            \centerline{\footnotesize (c) SRN\cite{Tao2018}}
	\end{minipage}
        \begin{minipage}{0.24\linewidth}
		\centering
		\includegraphics[width=1\linewidth]{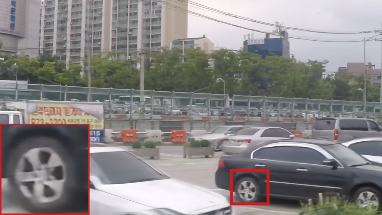}
            \centerline{\footnotesize (d) DMPHN\cite{Zhang2019}}
	\end{minipage}


        \begin{minipage}{0.24\linewidth}
		\centering
		\includegraphics[width=1\linewidth]{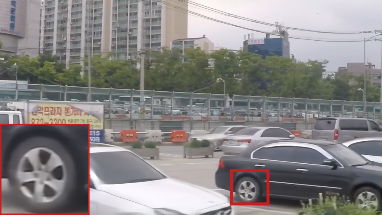}
            \centerline{\footnotesize (e) MIMO-UNet+\cite{Cho2021}}
	\end{minipage}
	\begin{minipage}{0.24\linewidth}
		\centering
		\includegraphics[width=1\linewidth]{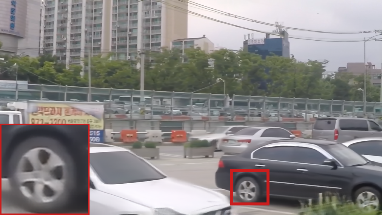}
            \centerline{\footnotesize (f) MPRNet\cite{Zamir2021}}
	\end{minipage}
        \begin{minipage}{0.24\linewidth}
		\centering
		\includegraphics[width=1\linewidth]{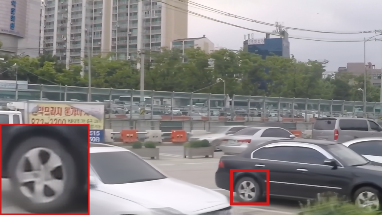}
            \centerline{\footnotesize (g) Restormer\cite{Zamir2022}}
	\end{minipage}
        \begin{minipage}{0.24\linewidth}
		\centering
		\includegraphics[width=1\linewidth]{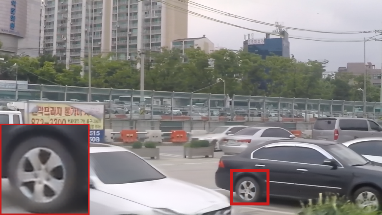}
            \centerline{\footnotesize (h) Ours}
	\end{minipage}

    \begin{minipage}{0.24\linewidth}
		\centering
		\includegraphics[width=1\linewidth]{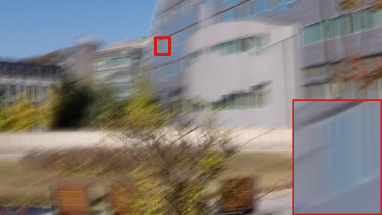}
            \centerline{\footnotesize (a) Blurred image}
	\end{minipage}
	\begin{minipage}{0.24\linewidth}
		\centering
		\includegraphics[width=1\linewidth]{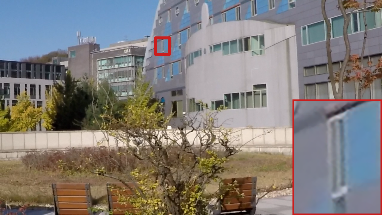}
            \centerline{\footnotesize (b) Sharp}
	\end{minipage}
        \begin{minipage}{0.24\linewidth}
		\centering
		\includegraphics[width=1\linewidth]{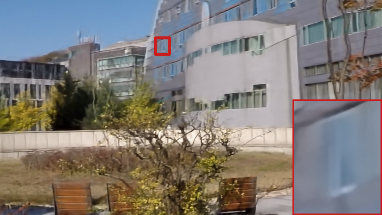}
            \centerline{\footnotesize (c) SRN\cite{Tao2018}}
	\end{minipage}
        \begin{minipage}{0.24\linewidth}
		\centering
		\includegraphics[width=1\linewidth]{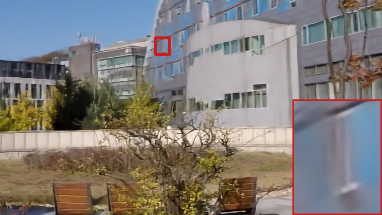}
            \centerline{\footnotesize (d) DMPHN\cite{Zhang2019}}
	\end{minipage}


        \begin{minipage}{0.24\linewidth}
		\centering
		\includegraphics[width=1\linewidth]{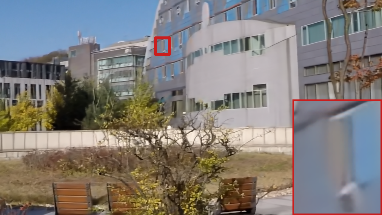}
            \centerline{\footnotesize (e) MIMO-UNet+\cite{Cho2021}}
	\end{minipage}
	\begin{minipage}{0.24\linewidth}
		\centering
		\includegraphics[width=1\linewidth]{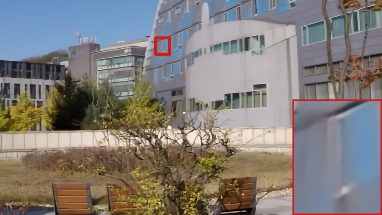}
            \centerline{\footnotesize (f) MPRNet\cite{Zamir2021}}
	\end{minipage}
        \begin{minipage}{0.24\linewidth}
		\centering
		\includegraphics[width=1\linewidth]{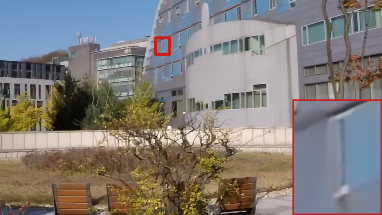}
            \centerline{\footnotesize (g) Restormer\cite{Zamir2022}}
	\end{minipage}
        \begin{minipage}{0.24\linewidth}
		\centering
		\includegraphics[width=1\linewidth]{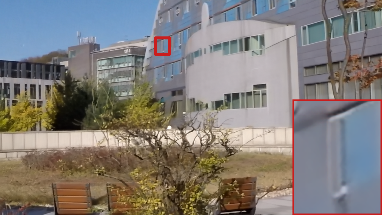}
            \centerline{\footnotesize (h) Ours}
	\end{minipage}

    \begin{minipage}{0.24\linewidth}
		\centering
		\includegraphics[width=1\linewidth]{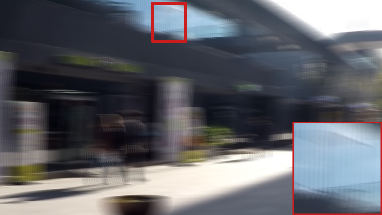}
            \centerline{\footnotesize (a) Blurred image}
	\end{minipage}
	\begin{minipage}{0.24\linewidth}
		\centering
		\includegraphics[width=1\linewidth]{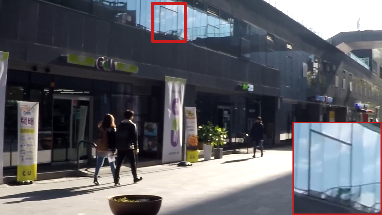}
            \centerline{\footnotesize (b) Sharp}
	\end{minipage}
        \begin{minipage}{0.24\linewidth}
		\centering
		\includegraphics[width=1\linewidth]{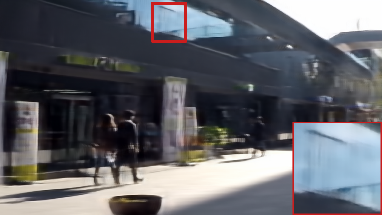}
            \centerline{\footnotesize (c) SRN\cite{Tao2018}}
	\end{minipage}
        \begin{minipage}{0.24\linewidth}
		\centering
		\includegraphics[width=1\linewidth]{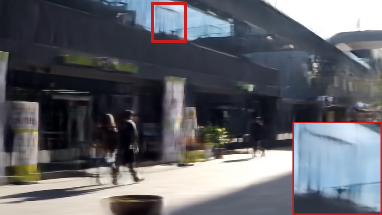}
            \centerline{\footnotesize (d) DMPHN\cite{Zhang2019}}
	\end{minipage}


        \begin{minipage}{0.24\linewidth}
		\centering
		\includegraphics[width=1\linewidth]{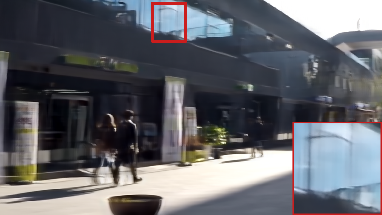}
            \centerline{\footnotesize (e) MIMO-UNet+\cite{Cho2021}}
	\end{minipage}
	\begin{minipage}{0.24\linewidth}
		\centering
		\includegraphics[width=1\linewidth]{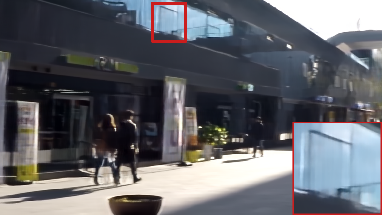}
            \centerline{\footnotesize (f) MPRNet\cite{Zamir2021}}
	\end{minipage}
        \begin{minipage}{0.24\linewidth}
		\centering
		\includegraphics[width=1\linewidth]{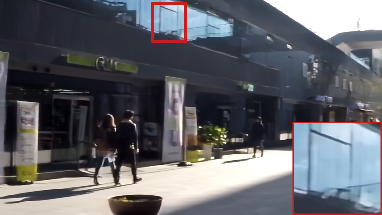}
            \centerline{\footnotesize (g) Restormer\cite{Zamir2022}}
	\end{minipage}
        \begin{minipage}{0.24\linewidth}
		\centering
		\includegraphics[width=1\linewidth]{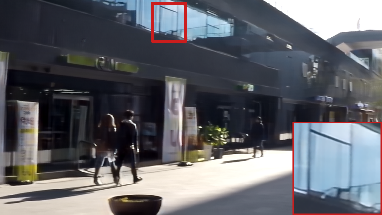}
            \centerline{\footnotesize (h) Ours}
	\end{minipage}

\caption{\footnotesize Visual comparisons of the different deblurring methods on the GoPro test dataset \cite{Nah2017}, our restorations are better than the state-of-the-art methods. From top left to bottom right: (a) blurry images,(b) Sharp images, and the restorations of (c) SRN\cite{Tao2018}, (d)DMPHN \cite{Zhang2019},(e) MIMO-UNet+\cite{Cho2021},(f) MPRNet\cite{Zamir2021},(g) Restormer\cite{Zamir2022} and (h) Ours, respectively}
\label{fig:4}
\end{figure}

\subsection{Loss Function}

The proposed method is trained by minimizing $L_1$ loss and frequency loss. Let $S \in 1, 2, 3$, $\hat{I}_{s}$, and $I_{gt}$ denote the stages in RDAFNet, the reconstructed image in the S-th stage and the ground-truth image, respectively. For any given stage S, we introduce two kinds of loss functions as follows. Following the recent work \cite{Cho2021}, we use $L_1$ loss to measure the distance between the output restored image and the ground-truth image:
\begin{equation}
     L_C=\left \| \hat{I}_{s}-I_{gt} \right \| _1
\end{equation}
The frequency loss, which measures the difference between the reconstructed image and the ground-truth sharp image in the frequency domain\cite{Cho2021}, is defined as: 
\begin{equation}
     L_f=\left \| \mathcal{F} \left (  \hat{I}_{s}\right ) -\mathcal{F} \left (  I_{gt}\right ) \right \| _1
\end{equation}
where $\mathcal{F}\left(\cdot \right)$ represents the fast Fourier transform(FFT) that transfers the image signal to the frequency domain.
The final loss function of network is the weighted sum of the above losses.
\begin{equation}
     L=\sum_{S=1}^{3} \left [L_{c}+\lambda L_{f} \right ] 
\end{equation}
The parameter $\lambda$ in Eq. (10) is set to 0.1 as in \cite{Cho2021}.

\begin{table}[!h]
\centering
\caption{\footnotesize{Deblurring results of various methods, our network is trained only on the GoPro dataset \cite{Nah2017} and directly evaluated on the HIDE dataset \cite{Zhang2018}. The best and sub-best results of the evaluation models are highlighted and underlined}}
\resizebox*{13.35cm}{82mm}{
\centering
\begin{tabular}{l|l|cc|cc|cc|c}
\toprule
\multirow{2}{*}{Model} &
  \multirow{2}{*}{Method} &
  \multicolumn{2}{c|}{GoPro\cite{Nah2017}} &
  \multicolumn{2}{c|}{HIDE\cite{Shen2019}} &
  \multicolumn{2}{c|}{Average} &
  Params \\
 &
   &
  PSNR$\uparrow$ &
  SSIM$\uparrow$ &
  PSNR$\uparrow$ &
  SSIM$\uparrow$ &
  PSNR$\uparrow$ &
  SSIM$\uparrow$ &
  (M) \\ \midrule \midrule
\multirow{16}{*}{\textbf{CNN-based model}} &
  Nah et al.\cite{Nah2017} &
  29.08 &
  0.914 &
  25.73 &
  0.874 &
  27.41 &
  0.894 &
  \multicolumn{1}{c}{11.7} \\
 &
  DeblurGAN\cite{Kupyn2018} &
  28.70 &
  0.858 &
  24.51 &
  0.871 &
  26.61 &
  0.865 &
  - \\
 &
  Zhang et al.\cite{Zhang2018a} &
  29.19 &
  0.931 &
  - &
  - &
  - &
  - &
  9.2 \\
 &
  DeburGAN-v2\cite{Kupyn2019} &
  29.55 &
  0.934 &
  26.61 &
  0.875 &
  28.08 &
  0.905 &
  60.9 \\
 &
  SRN\cite{Tao2018} &
  30.26 &
  0.934 &
  28.36 &
  0.915 &
  29.31 &
  0.925 &
  6.8 \\
 &
  Gao et al.\cite{Gao2019} &
  30.90 &
  0.935 &
  29.11 &
  0.913 &
  30.01 &
  0.924 &
  11.3 \\
 &
  DBGAN\cite{Zhang2020a} &
  31.10 &
  0.943 &
  28.94 &
  0.915 &
  30.02 &
  0.929 &
  11.6 \\
 &
  DGN\cite{Li2020} &
  30.49 &
  0.938 &
  - &
  - &
  - &
  - &
  11.32 \\
 &
  MT-RNN \cite{Chan1998}&
  31.15 &
  0.945 &
  29.15 &
  0.918 &
  30.15 &
  0.932 &
  2.6 \\
 &
  DMPHN\cite{Zhang2019} &
  31.20 &
  0.940 &
  29.09 &
  0.924 &
  30.15 &
  0.932 &
  21.7 \\
 &
  SDWNet\cite{Zou2021} &
  31.26 &
  0.966 &
  28.99 &
  0.957 &
  30.13 &
  0.962 &
  7.2 \\
 &
  Suin et al.\cite{Suin2020} &
  31.85 &
  0.948 &
  29.98 &
  0.930 &
  30.92 &
  0.939 &
  23.0 \\
 &
  RADN \cite{Purohit2020}&
  31.85 &
  0.953 &
  - &
  - &
  - &
  - &
  - \\
 &
  MIMO-UNet+\cite{Cho2021} &
  32.45 &
  0.957 &
  29.99 &
  0.930 &
  31.22 &
  0.944 &
  16.1 \\
 &
  MPRNet\cite{Zamir2021} &
  32.66 &
  0.959 &
  30.96 &
  0.939 &
  31.81 &
  0.949 &
  20.1 \\
 &
  HINet\cite{Chen2021} &
  32.71 &
  0.959 &
  30.32 &
  0.932 &
  31.52 &
  0.946 &
  \multicolumn{1}{c}{88.7} \\ \midrule
\multirow{2}{*}{\textbf{\begin{tabular}[c]{@{}l@{}}Transformer-based \\ models\end{tabular}}} &
  Pretrained-IPT\cite{Chen2021} &
  32.52 &
  - &
  - &
  - &
  - &
  - &
  \multicolumn{1}{c}{114} \\
 &
  Restormer\cite{Zamir2022} &
  {\ul 32.92} &
  {\ul 0.961} &
  {\ul 31.22} &
  \textbf{0.942} &
  {\ul 32.07} &
  \textbf{0.952} &
  \multicolumn{1}{c}{26.12} \\ \midrule
\multirow{2}{*}{\textbf{\begin{tabular}[c]{@{}l@{}}Our model(CNN-\\ based)\end{tabular}}} &
  Single-stage RDAFNet &
  32.24 &
  0.955 &
  30.57 &
  0.934 &
  31.41 &
  0.945 &
  \multicolumn{1}{c}{14.2} \\
 &
  Multi-stage RDAFNet &
  \textbf{33.06} &
  \textbf{0.961} &
  \textbf{31.28} &
  {\ul 0.941} &
  \textbf{32.17} &
  {\ul 0.951} &
  \multicolumn{1}{c}{28.97} \\ \bottomrule
\end{tabular}
}
\label{tab:table1}
\end{table}

\section{Experiments}\label{sec:exp}
\subsection{Implementation details}

We use a GoPro dataset \cite{Nah2017} that has 3214 pairs of blurred images and sharp images of size $720 \times 1280$, where 2103 pairs are used for training and the remaining 1111 pairs are used for testing. The blurred images of the GoPro dataset are generated by successively integrating sharp frames and the corresponding ground truth sharp images that are captured by a high-speed camera. To demonstrate our model’s generalizability, we directly apply our GoPro \cite{Nah2017} training model on the HIDE \cite{Shen2019} and some real-world blurred images provided in \cite{Lai2016}. 

To train the network, we used the Adam solver to optimize the network parameters and set the $\beta_1$,$\beta_2$ to 0.9, 0.99. Our network is trained for 3000 epochs. The initial learning rates is set to be $10^{-4}$. After 300 epochs, the learning rate starts to linearly decrease to by
\begin{equation}
     L_r=10^{-4}-\frac{10^{-4}-10^{-7}}{2700} \left (M-300  \right ) 
\end{equation}
where $M$ denotes the number of train epoch.

\begin{figure}[!hbt]
\centering
	\begin{minipage}{0.24\linewidth}
		\centering
		\includegraphics[width=1\linewidth]{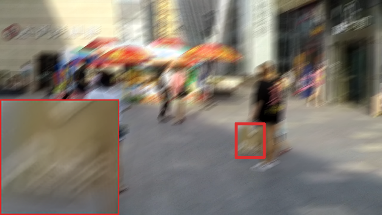}
            \centerline{\footnotesize (a) Blurred image}
	\end{minipage}
	\begin{minipage}{0.24\linewidth}
		\centering
		\includegraphics[width=1\linewidth]{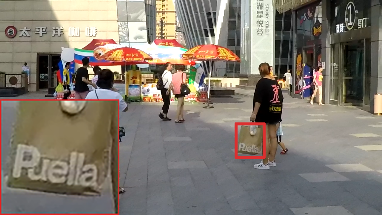}
            \centerline{\footnotesize (b) Sharp}
	\end{minipage}
        \begin{minipage}{0.24\linewidth}
		\centering
		\includegraphics[width=1\linewidth]{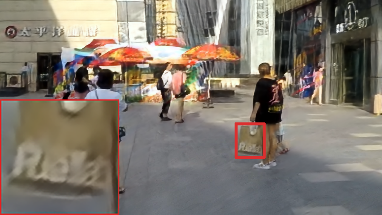}
            \centerline{\footnotesize (c) SRN\cite{Tao2018}}
	\end{minipage}
        \begin{minipage}{0.24\linewidth}
		\centering
		\includegraphics[width=1\linewidth]{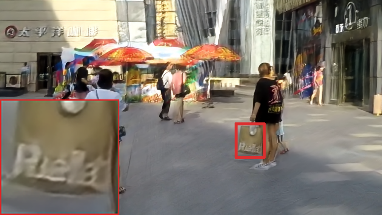}
            \centerline{\footnotesize (d) DMPHN\cite{Zhang2019}}
	\end{minipage}


        \begin{minipage}{0.24\linewidth}
		\centering
		\includegraphics[width=1\linewidth]{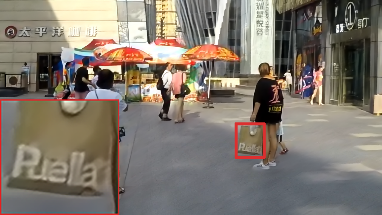}
            \centerline{\footnotesize (e) MIMO-UNet+\cite{Cho2021}}
	\end{minipage}
	\begin{minipage}{0.24\linewidth}
		\centering
		\includegraphics[width=1\linewidth]{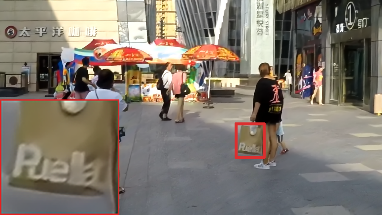}
            \centerline{\footnotesize (f) MPRNet\cite{Zamir2021}}
	\end{minipage}
        \begin{minipage}{0.24\linewidth}
		\centering
		\includegraphics[width=1\linewidth]{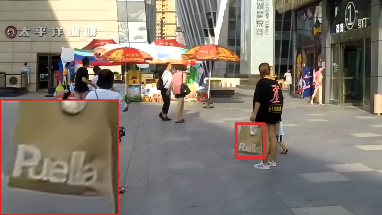}
            \centerline{\footnotesize (g) Restormer\cite{Zamir2022}}
	\end{minipage}
        \begin{minipage}{0.24\linewidth}
		\centering
		\includegraphics[width=1\linewidth]{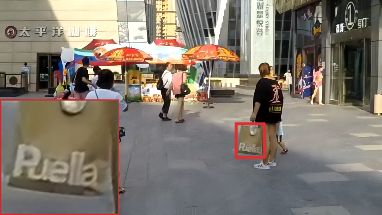}
            \centerline{\footnotesize (h) Ours}
	\end{minipage}

    \begin{minipage}{0.24\linewidth}
		\centering
		\includegraphics[width=1\linewidth]{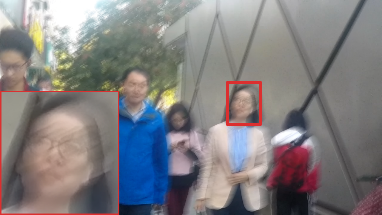}
            \centerline{\footnotesize (a) Blurred image}
	\end{minipage}
	\begin{minipage}{0.24\linewidth}
		\centering
		\includegraphics[width=1\linewidth]{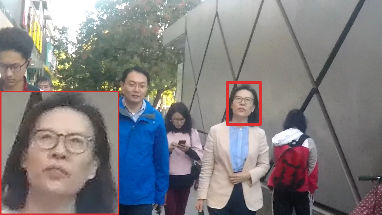}
            \centerline{\footnotesize (b) Sharp}
	\end{minipage}
        \begin{minipage}{0.24\linewidth}
		\centering
		\includegraphics[width=1\linewidth]{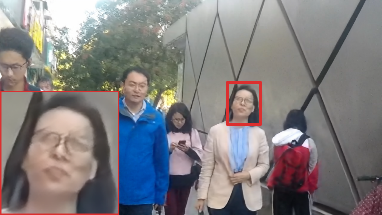}
            \centerline{\footnotesize (c) SRN\cite{Tao2018}}
	\end{minipage}
        \begin{minipage}{0.24\linewidth}
		\centering
		\includegraphics[width=1\linewidth]{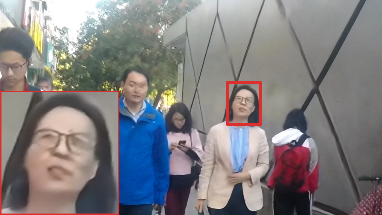}
            \centerline{\footnotesize (d) DMPHN\cite{Zhang2019}}
	\end{minipage}


        \begin{minipage}{0.24\linewidth}
		\centering
		\includegraphics[width=1\linewidth]{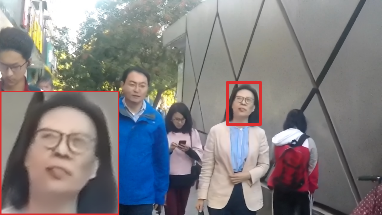}
            \centerline{\footnotesize (e) MIMO-UNet+\cite{Cho2021}}
	\end{minipage}
	\begin{minipage}{0.24\linewidth}
		\centering
		\includegraphics[width=1\linewidth]{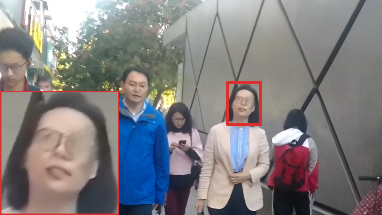}
            \centerline{\footnotesize (f) MPRNet\cite{Zamir2021}}
	\end{minipage}
        \begin{minipage}{0.24\linewidth}
		\centering
		\includegraphics[width=1\linewidth]{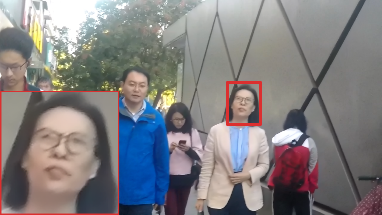}
            \centerline{\footnotesize (g) Restormer\cite{Zamir2022}}
	\end{minipage}
        \begin{minipage}{0.24\linewidth}
		\centering
		\includegraphics[width=1\linewidth]{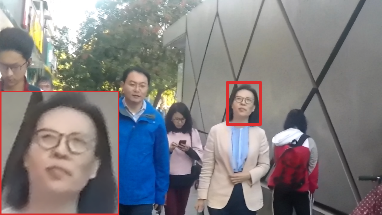}
            \centerline{\footnotesize (h) Ours}
	\end{minipage}

    \begin{minipage}{0.24\linewidth}
		\centering
		\includegraphics[width=1\linewidth]{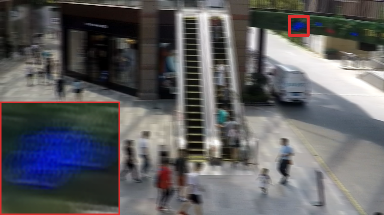}
            \centerline{\footnotesize (a) Blurred image}
	\end{minipage}
	\begin{minipage}{0.24\linewidth}
		\centering
		\includegraphics[width=1\linewidth]{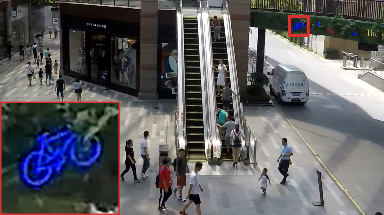}
            \centerline{\footnotesize (b) Sharp}
	\end{minipage}
        \begin{minipage}{0.24\linewidth}
		\centering
		\includegraphics[width=1\linewidth]{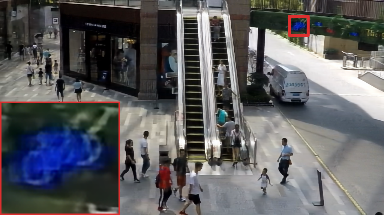}
            \centerline{\footnotesize (c) SRN\cite{Tao2018}}
	\end{minipage}
        \begin{minipage}{0.24\linewidth}
		\centering
		\includegraphics[width=1\linewidth]{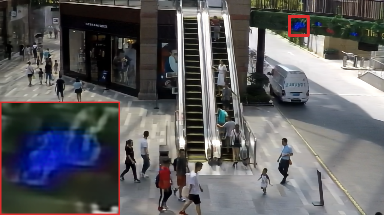}
            \centerline{\footnotesize (d) DMPHN\cite{Zhang2019}}
	\end{minipage}


        \begin{minipage}{0.24\linewidth}
		\centering
		\includegraphics[width=1\linewidth]{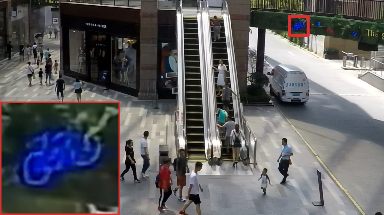}
            \centerline{\footnotesize (e) MIMO-UNet+\cite{Cho2021}}
	\end{minipage}
	\begin{minipage}{0.24\linewidth}
		\centering
		\includegraphics[width=1\linewidth]{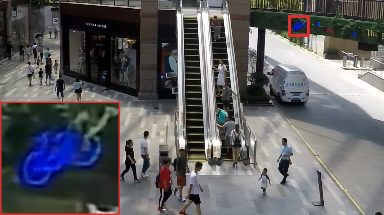}
            \centerline{\footnotesize (f) MPRNet\cite{Zamir2021}}
	\end{minipage}
        \begin{minipage}{0.24\linewidth}
		\centering
		\includegraphics[width=1\linewidth]{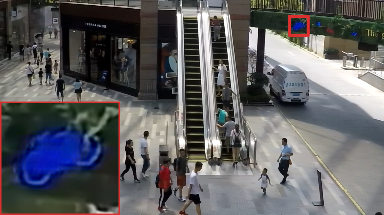}
            \centerline{\footnotesize (g) Restormer\cite{Zamir2022}}
	\end{minipage}
        \begin{minipage}{0.24\linewidth}
		\centering
		\includegraphics[width=1\linewidth]{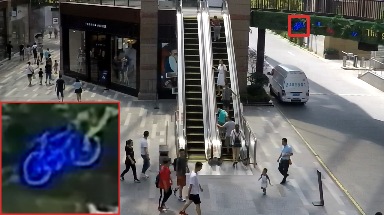}
            \centerline{\footnotesize (h) Ours}
	\end{minipage}

\caption{\footnotesize Visual comparisons of the different deblurring methods on the HIDE test dataset \cite{Zhang2020a}, our network is trained only on the GoPro dataset [11], and our restorations are better than the state-of-the-art methods. From top left to bottom right: (a) blurry images,(b) Sharp images, and the restorations of (c) SRN\cite{Tao2018}, (d) DMPHN\cite{Zhang2019}, (e) MIMO-UNet+\cite{Cho2021}, (f) MPRNet\cite{Zamir2021}, (g) Restormer\cite{Zamir2022} and (h) Ours, respectively.}
\label{fig:5}
\end{figure}

We randomly cropped a $256 \times 256$ region from the blurred image and used the corresponding sharp image as the input to the network. All input images were normalized to a range between -1 and 1. For data augmentation, we randomly applied different combinations of vertical and horizontal flip and rotate operations to the training data. All experiments were implemented on a PC with an Intel i9-10101 CPU and an NVIDIA RTX 3090 GPU. 

\begin{figure}[!h]
\centering
	\begin{minipage}{0.24\linewidth}
		\centering
		\includegraphics[width=1\linewidth]{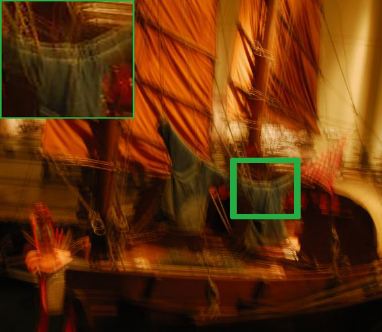}
            \centerline{\footnotesize (a) Blurred image}
	\end{minipage}
	\begin{minipage}{0.24\linewidth}
		\centering
		\includegraphics[width=1\linewidth]{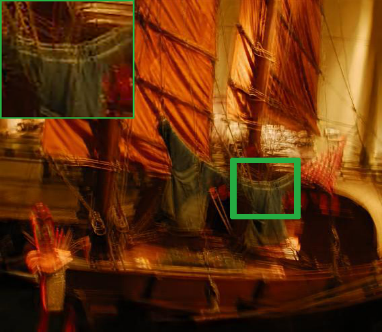}
            \centerline{(b) SRN}
	\end{minipage}
        \begin{minipage}{0.24\linewidth}
		\centering
		\includegraphics[width=1\linewidth]{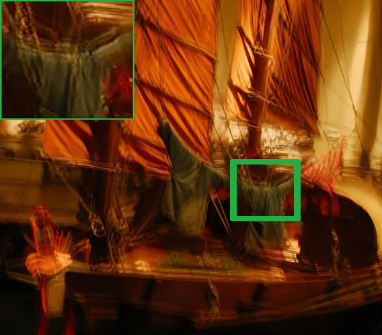}
            \centerline{(c) Deblurgan-v2}
	\end{minipage}
        \begin{minipage}{0.24\linewidth}
		\centering
		\includegraphics[width=1\linewidth]{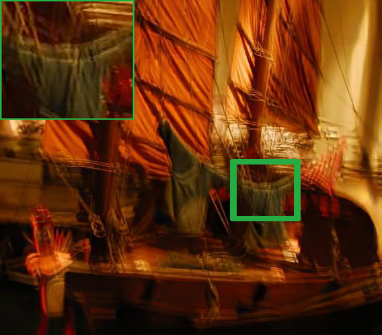}
            \centerline{\footnotesize (d) DMPHN\cite{Zhang2019}}
	\end{minipage}


        \begin{minipage}{0.24\linewidth}
		\centering
		\includegraphics[width=1\linewidth]{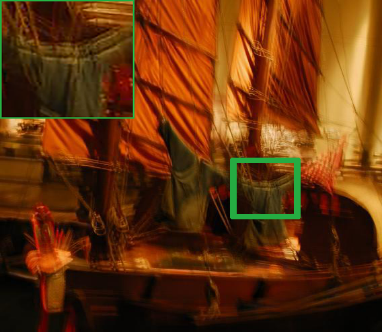}
            \centerline{\footnotesize (e) MIMO-UNet+\cite{Cho2021}}
	\end{minipage}
	\begin{minipage}{0.24\linewidth}
		\centering
		\includegraphics[width=1\linewidth]{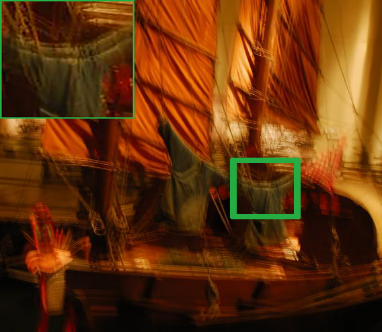}
            \centerline{\footnotesize (f) MPRNet\cite{Zamir2021}}
	\end{minipage}
        \begin{minipage}{0.24\linewidth}
		\centering
		\includegraphics[width=1\linewidth]{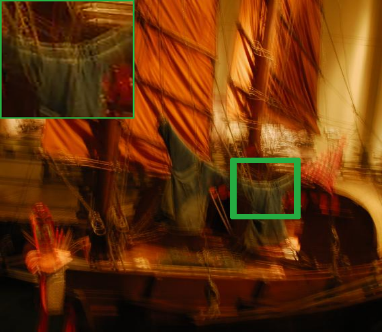}
            \centerline{\footnotesize (g) Restormer\cite{Zamir2022}}
	\end{minipage}
        \begin{minipage}{0.24\linewidth}
		\centering
		\includegraphics[width=1\linewidth]{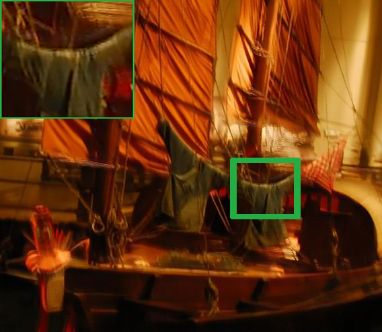}
            \centerline{\footnotesize (h) Ours}
	\end{minipage}

    \begin{minipage}{0.24\linewidth}
		\centering
		\includegraphics[width=1\linewidth]{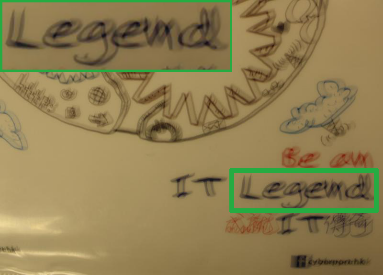}
            \centerline{\footnotesize (a) Blurred image}
	\end{minipage}
	\begin{minipage}{0.24\linewidth}
		\centering
		\includegraphics[width=1\linewidth]{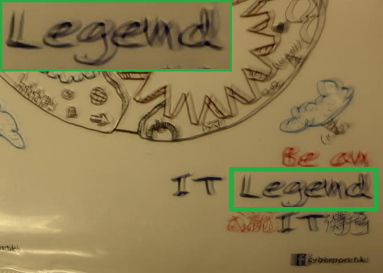}
            \centerline{(b) SRN}
	\end{minipage}
        \begin{minipage}{0.24\linewidth}
		\centering
		\includegraphics[width=1\linewidth]{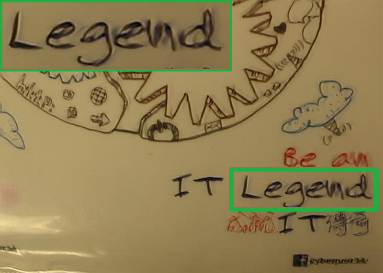}
            \centerline{(c) Deblurgan-v2}
	\end{minipage}
        \begin{minipage}{0.24\linewidth}
		\centering
		\includegraphics[width=1\linewidth]{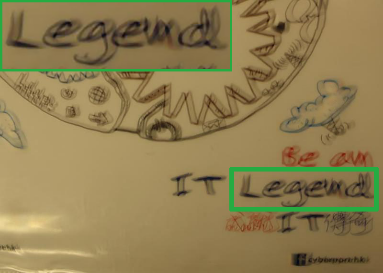}
            \centerline{\footnotesize (d) DMPHN\cite{Zhang2019}}
	\end{minipage}


        \begin{minipage}{0.24\linewidth}
		\centering
		\includegraphics[width=1\linewidth]{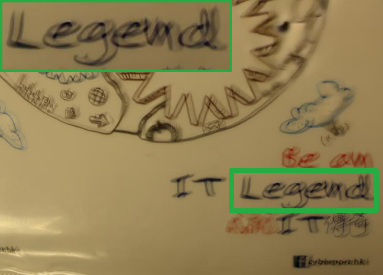}
            \centerline{\footnotesize (e) MIMO-UNet+\cite{Cho2021}}
	\end{minipage}
	\begin{minipage}{0.24\linewidth}
		\centering
		\includegraphics[width=1\linewidth]{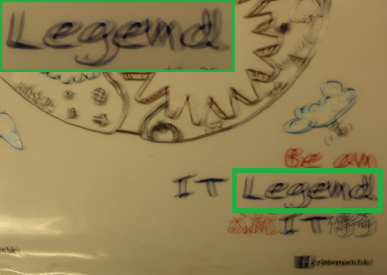}
            \centerline{\footnotesize (f) MPRNet\cite{Zamir2021}}
	\end{minipage}
        \begin{minipage}{0.24\linewidth}
		\centering
		\includegraphics[width=1\linewidth]{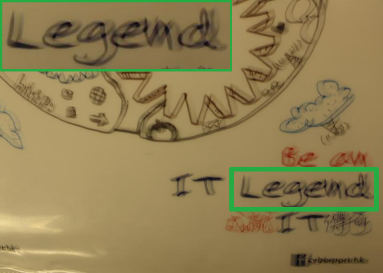}
            \centerline{\footnotesize (g) Restormer\cite{Zamir2022}}
	\end{minipage}
        \begin{minipage}{0.24\linewidth}
		\centering
		\includegraphics[width=1\linewidth]{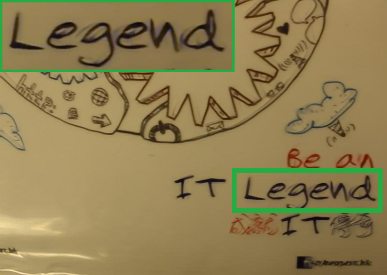}
            \centerline{\footnotesize (h) Ours}
	\end{minipage}

\caption{\footnotesize Visual comparisons on real-word blur images from \cite{Lai2016}, our network is trained only on the GoPro dataset \cite{Nah2017}, and our restorations are better than the state-of-the-art methods. From top left to bottom right: (a) blurry images and the restorations of (b) SRN \cite{Tao2018}, (c) Deblurgan-v2 \cite{Zhang2018a} , (d) DMPHN\cite{Zhang2019}, (e) MIMO-UNet+\cite{Cho2021}, (f) MPRNet\cite{Zamir2021}, (g) Restormer\cite{Zamir2022} and  (h) Ours, respectively.}
\label{fig:6}
\end{figure}

In our experiments, the single-stage RDAFNet stacks 16 RDAFBs, and the multi-stage RDAFNet contains three subnetworks, each of which employs 10 RDAFBs, 10 RDAFBs and 12 RDAFBs respectively. In each RDAFB, we set the number of convolution layers to 4, and each convolution layer has 128 filters. For testing, we using the same method as use in recent state-of-the-art models MPRNet \cite{Zamir2021} and Restormer \cite{Zamir2022}, which first saves the deblurring results of network as png images, and then the peak signal to noise ratio (PSNR) and structural similarity (SSIM) are computed by matlab.  

\subsection{Experimental results}

1) Quantitative Analysis: We compared the proposed method with the existing state-of-the-art deblurring models \cite{Nah2017}, \cite{Kupyn2018}, \cite{Zhang2018a}, \cite{Kupyn2019}, \cite{Tao2018}, \cite{Gao2019}, \cite{Zhang2020a}, \cite{Li2020}, \cite{Chan1998}, \cite{Zhang2019}, \cite{Zou2021}, \cite{Suin2020}, \cite{Purohit2020}, \cite{Cho2021}, \cite{Zamir2021}, \cite{chen2021hinet}, \cite{Chen2021} ,\cite{Zamir2022}. Table \ref{tab:table1} lists the PSNR and SSIM values tested on the GoPro and HIDE datasets from the above methods. From Table \ref{tab:table1}, we can see that our multi-sage RDAFNet outperforms the all the other compared models with the best PSNR. Compared with the previous best performing CNN-based model HINet \cite{Xu2013}, our multi-stage RDAFNet uses about only one-third of the parameters of HINet and achieves a 0.35dB improvement in PSNR on GoPro dataset. 

Table \ref{tab:table1} also shows that only our method achieves the best PSNR on the GoPro dataset, as well as the best PSNR on the HIDE dataset. It is worth noting that we tested the HIDE dataset using the model trained only on the GoPro dataset, which demonstrated the strong generalization abilities of the proposed model. Moreover, our multi-stage RDAFNet outperforms the transformer model IPT \cite{Chen2021} by 0.54dB in PSNR on GoPro dataset. Compared to a recent transformer method Restormer \cite{Zamir2022}, our method also achieve comparable results.

2) Qualitative Analysis: Figure \ref{fig:4} and Figure \ref{fig:5} show the visual comparisons on the GoPro and HIDE datasets, respectively. We compared proposed method with five state-of-the-art methods, SRN \cite{Tao2018}, DMPHN \cite{Zhang2019}, MIMO-UNet+ \cite{Cho2021}, MPRNet \cite{Zamir2021} and Restormer \cite{Zamir2022}. To fully show the different methods’ deblurring results, we focused on the details. As shown in Fig. \ref{fig:4}, compared with the results of previous methods, deblurring is not good on the text regions and the contours, and our RDAFNet recovers these regions better. In Figure \ref{fig:5}, the results of the old methods still have some blur on the faces, but RDAFNet can accurately restore a clear image. Overall, the image restored by our model has clearer edges and richer details than the other methods and is very close to the ground-truth image. 


To further evaluate the generalization performance of our RDAFNet in real scenarios, we also applied the GoPro trained model on real-world blurry images. Figure \ref{fig:6} shows the visual comparisons on real-world blur images provided in \cite{Lai2016}. Compared with the six state-of-the-art methods, SRN \cite{Tao2018}, Deblurgan-v2 \cite{Kupyn2019}, DMPHN \cite{Zhang2019}, MIMO-UNet+ \cite{Cho2021}, MPRNet \cite{Dai2019} and Restormer \cite{Zamir2022}, our method is still able to restore sharper and more natural images.

\begin{table}[]
\caption{\footnotesize{Ablation experiments with different combinations of AM, AFM and LRL, the “$*$” indicates that the model is not convergent during training.}}
\tabcolsep=0.5cm
\resizebox{\textwidth}{17.5mm}{
\begin{tabular}{c|l|ccc|ccc}
\toprule
\multicolumn{1}{l|}{}                                                              & Method             & AM & AFM & LRL & PSNR  & SSIM  & Params \\ \midrule \midrule
\multirow{5}{*}{\begin{tabular}[c]{@{}c@{}}Without CCLAT\\ Mechanism\end{tabular}} & AM(0)AFM(0)LRL(0)          &  \ding{55}  &  \ding{55}   &  \ding{55}  & 29.12 & 0.915 & 9.93M      \\
 & AM(1)AFM(0)LRL(0)*  & \checkmark & \ding{55} & \ding{55} & -     & -     & 10.99M \\
 & AM(1)AFM(1)LRL(0)*  & \checkmark & \checkmark & \ding{55} & -     & -     & 14.20M \\
 & AM(0)AFM(0)LRL(1)   & \ding{55} & \ding{55} & \checkmark & 31.60 & 0.946 & 9.93M  \\
 & AM(1)AFM(0)LRL(1)   & \checkmark & \ding{55} & \checkmark & 31.68 & 0.947 & 10.99M \\ \midrule
\begin{tabular}[c]{@{}c@{}}With CCLAT\\ Mechanism\end{tabular}                     & AM(1)AFM(1)LRL(1)    & \checkmark   & \checkmark    & \checkmark    & 32.24 & 0.955 & 14.20M     \\ \midrule \midrule
\multicolumn{1}{l|}{} &
  Baseline &
  \multicolumn{3}{l|}{UNet with Resblocks\cite{Nah2017}} &
  \multicolumn{1}{c}{31.77} &
  \multicolumn{1}{c}{0.951} &
  \multicolumn{1}{c}{14.4M} \\ \bottomrule
\end{tabular}
}
\label{tab:table2}
\end{table}

\begin{table}[]\footnotesize
\centering
\caption{\footnotesize{Efficiency comparison on GoPro dataset \cite{Nah2017}. FLOPs are calculated with the input size of 256 × 256.}}
\resizebox*{140mm}{30mm}{
\begin{tabular}{l|cccc|cc}
\toprule
\multirow{2}{*}{Method} & 
\multirow{2}{*}{\begin{tabular}[c]{@{}c@{}}DBGAN\\ \cite{Zhang2019a}\end{tabular}} & 
\multirow{2}{*}{\begin{tabular}[c]{@{}c@{}}DMPHN\\ \cite{Zhang2019}\end{tabular}} & \multirow{2}{*}{\begin{tabular}[c]{@{}c@{}}Suin et al.\\ \cite{Suin2020}\end{tabular}} & \multirow{2}{*}{\begin{tabular}[c]{@{}c@{}}MPRNet\\ \cite{Zamir2021}\end{tabular}} & \multicolumn{2}{c}{RDAFNet} \\ \cline{6-7} 
          &       &        &        &       & \multicolumn{1}{c}{Single-stage} & Multi-Stage \\ \midrule \midrule
Params(M) & 11.6  & 21.7   & 23.0   & 21.0  & \multicolumn{1}{c}{14.20}        & 28.97       \\
Flops(G)  & 660.2 & 678.56 & 536.74 & 660.2 & \multicolumn{1}{c}{241.52}       & 502.62      \\
PSNR      & 31.10 & 31.20  & 31.85  & 32.66 & \multicolumn{1}{c}{32.24}        & 33.06       \\
SSIM      & 0.942 & 0.940  & 0.948  & 0.959 & \multicolumn{1}{c}{0.955}        & 0.961       \\ \bottomrule
\end{tabular}
}
\label{tab:table3}
\end{table}

\subsection{Effectiveness of Continuous Cross-Layer Attention Transmission Mechanism }
The fundamental block in our networks is the RDAFB, which is designed based on the continuous cross-layer attention transmission (CCLAT) mechanism. The RDAFB consists of attentional module (AM), attentional fusion module (AFM) and local residual learning (LRL), thus forming a continuous cross-layer attentional transmission mechanism. To comprehensively verify the effectiveness of continuous cross-layer attention transmission mechanism, the RDAFB is evaluated from multiple perspectives. 

We use the single-stage RDAFNet structure for the ablation experiments, and all evaluations are performed on the GoPro dataset \cite{Nah2017}. A series of ablation studies are provided in Table \ref{tab:table2}. Since the AFM design is based on AM, there are a total of six different combinations of ablation studies. We denote the baseline as AM$\left(0\right)$AFM$\left(0\right)$LRL$\left(0\right)$, which is obtained without AM, AFM and LRL, and the PSNR value test on the baseline is 29.12dB. 

From Table \ref{tab:table2}, we observe that the performance of the three networks without LRL is very poor, and even the training of AM$\left(1\right)$AFM$\left(0\right)$LRL$\left(0\right)$ and AM$\left(1\right)$AFM$\left(1\right)$LRL$\left(0\right)$ are not convergent. When we add LRL to the baseline, the PSNR of AM$\left(0\right)$AFM$\left(0\right)$LRL$\left(1\right)$ increases by 2.48dB compared to the baseline, which further demonstrates that these networks need LRL for stable training, so LRL cannot be removed.

Then we add AM to AM$\left(0\right)$AFM$\left(0\right)$LRL$\left(1\right)$, resulting in AM(1)AFM(0)L-RL(1), which brings a 0.08dB gain in the PSNR values. This indicates that the attention module we use here can bring performance improvement.

Finally we used the CCLAT mechanism-based RDAFB that contains all the components (denoted as AM$\left(1\right)$AFM$\left(1\right)$LRL$\left(1\right)$), and the PSNR value increased by 0.56 dB compared to the best performance model without CCLAT mechanism (denote as AM$\left(1\right)$AFM$\left(0\right)$LRL$\left(1\right)$). Furthermore, we compare AM(1)AFM(1)LRL(1) with the baseline, which serve ResBlocks\cite{Nah2017} as building blocks. Table \ref{tab:table2} shows that CCLAT mechanism-based AM(1)AFM(1)LR-L(1) provides favorable gain of 0.48 dB over the baseline. This result demonstrates the effectiveness of continuous cross-layer attention transmission mechanism.

\begin{figure}
\centering
	\begin{minipage}{0.24\linewidth}
		\centering
		\includegraphics[width=1\linewidth]{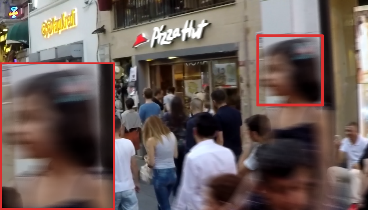}
	\end{minipage}
	\begin{minipage}{0.24\linewidth}
		\centering
		\includegraphics[width=1\linewidth]{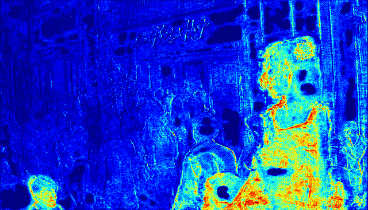}
	\end{minipage}
        \begin{minipage}{0.24\linewidth}
		\centering
		\includegraphics[width=1\linewidth]{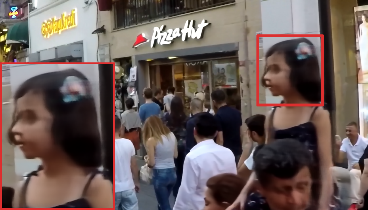}
	\end{minipage}
        \begin{minipage}{0.24\linewidth}
		\centering
		\includegraphics[width=1\linewidth]{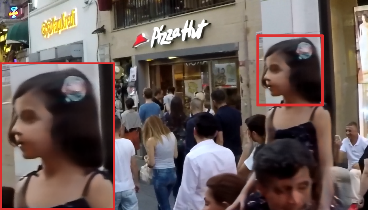}
	\end{minipage}

\vspace{5.5pt}

        \begin{minipage}{0.24\linewidth}
		\centering
		\includegraphics[width=1\linewidth]{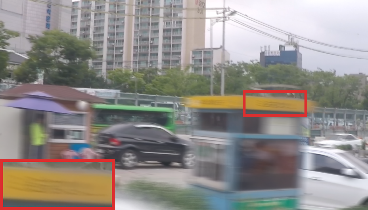}
	\end{minipage}
	\begin{minipage}{0.24\linewidth}
		\centering
		\includegraphics[width=1\linewidth]{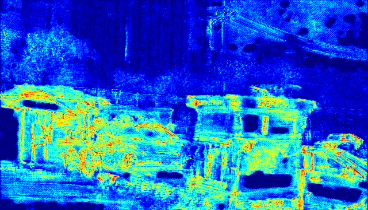}
	\end{minipage}
        \begin{minipage}{0.24\linewidth}
		\centering
		\includegraphics[width=1\linewidth]{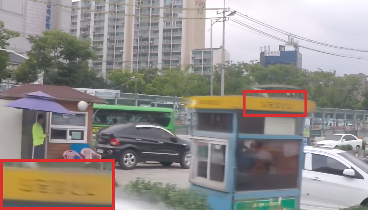}
	\end{minipage}
        \begin{minipage}{0.24\linewidth}
		\centering
		\includegraphics[width=1\linewidth]{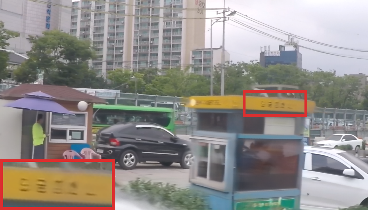}
	\end{minipage}

\vspace{5.5pt}

    \begin{minipage}{0.24\linewidth}
		\centering
		\includegraphics[width=1\linewidth]{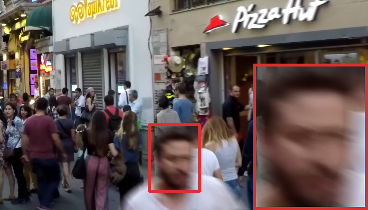}
	\end{minipage}
	\begin{minipage}{0.24\linewidth}
		\centering
		\includegraphics[width=1\linewidth]{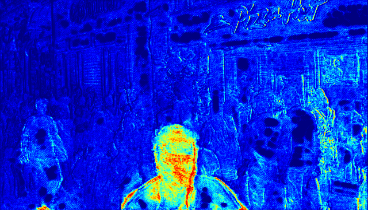}
	\end{minipage}
        \begin{minipage}{0.24\linewidth}
		\centering
		\includegraphics[width=1\linewidth]{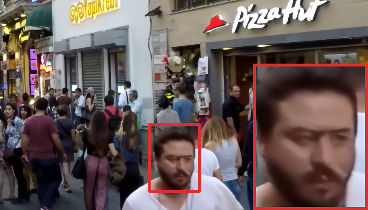}
	\end{minipage}
        \begin{minipage}{0.24\linewidth}
		\centering
		\includegraphics[width=1\linewidth]{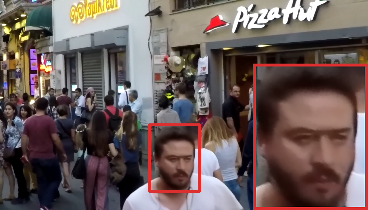}
	\end{minipage}
 

    \begin{minipage}{0.24\linewidth}
		\centering
  \vspace{5.5pt}
		\includegraphics[width=1\linewidth]{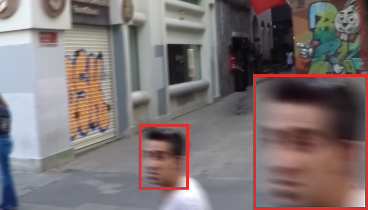}
            \centerline{\footnotesize \makecell{(a) Non-uniform \\ blurred images}}
            \vspace{-5pt}
	\end{minipage}
	\begin{minipage}{0.24\linewidth}
		\centering
		\includegraphics[width=1\linewidth]{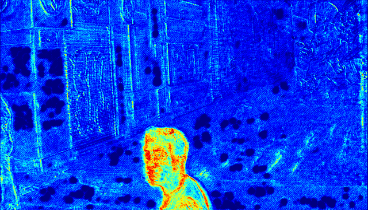}
            \centerline{\footnotesize (b) Attention maps}
	\end{minipage}
        \begin{minipage}{0.24\linewidth}
		\centering
		\includegraphics[width=1\linewidth]{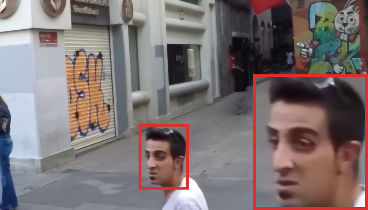}
            \centerline{\footnotesize (c) MPRNet \cite{Zamir2021}}
	\end{minipage}
        \begin{minipage}{0.24\linewidth}
		\centering
		\includegraphics[width=1\linewidth]{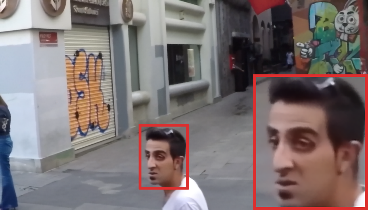}
            \centerline{\footnotesize (d) Ours}
	\end{minipage}

\caption{\footnotesize Visual comparisons on the non-uniform blurred images caused by camera movement or object movement, From left to right: (a) non-uniform blurry images, (b) one of the attention maps corresponding to each input blurred image, which come from the CCLAT mechanism-based RDAFBs, red indicates higher values in the attention map, while blue indicates lower values, (c) the restorations of MPRNet \cite{Zamir2021}, (d) Ours deblurring results}
\label{fig:7}
\end{figure}

\subsection{FLOPs and Model Size.}

Since the consumption of resources can reflect the efficiency of the model, we analyse the parameters and computational complexity of different models in this subsection. Considering that the computation time is highly dependent on the hardware conditions and environment, we compare the FLOPs of our model with those of previous models. The experimental results are provided in Table \ref{tab:table3}. Notably, our single-stage model RDAFNet is light, fast and produces better results than other complex algorithms such as DHPHN \cite{Zhang2019} and Suin et al. \cite{Suin2020}. Similarly, when compared to the recent method MPRNet \cite{Zamir2021}, our multi-stage RDAFNet achieves performance gains with higher computational efficiency.

\section{Analysis and discussion}
\subsection{Non-Uniform Deblurring}

Different from the motion blur in static scene, the dynamic scene blur is caused by camera movement, rigid or non-rigid object movement, and the variation of scene depth, which is non-uniform. To further demonstrate the effectiveness of the proposed method, we show the deblurring results on non-uniform blur images. The first column of Figure \ref{fig:7} shows the input images that suffer from complex non-uniform blur. The second column of Figure \ref{fig:7} shows one of the attention maps corresponding to each input blurred image. We can clearly see that the estimated weight of the attention map is highly correlated to the blurred regions, which brings performance improvement benefits. The fourth column of Figure \ref{fig:7} provides the deblurring results of our method.  Compared with the MPRNet \cite{Zamir2021}, the faces and texts restored by our method are clearer and has richer details. From the observations, we conjecture that the CCLAT mechanism-based RDAFBs enables the network to focus on the information that is most useful to the non-uniform deblurring in dynamic scenes.

\begin{table}[]\footnotesize
\centering
\caption{\footnotesize{Performance comparison of different blocks. All models have same number of convolution layers and similar model size}}
\begin{tabular}{l|l|ccc}
\toprule
 & Method & PSNR  & SSIM  & Params \\ \midrule \midrule
\multirow{2}{*}{\begin{tabular}[c]{@{}l@{}}32-Convolutional\\ Layers\end{tabular}} & RDB\cite{Zhang2020} & 31.02 & 0.934 & 7.34M  \\
 & RDAFB  & 31.64 & 0.947 & 7.34M      \\ \midrule
\multirow{2}{*}{\begin{tabular}[c]{@{}l@{}}64-Convolutional\\ Layers\end{tabular}} & RDB\cite{Zhang2020} & 31.79 & 0.948 & 14.20M \\
 & RDAFB  & 32.24 & 0.955 & 14.20M     \\ \bottomrule
\end{tabular}
\label{tab:table4}
\end{table}

\begin{table}\footnotesize
\centering
\caption{\footnotesize{Comparison with stage-wise model of RDAFNet on GoPro dataset \cite{Nah2017}, N represents the number of RDAFBs.}}
\begin{tabular}{l|cccc|cc}
\toprule
\multirow{2}{*}{RDAFNet} &
  \multicolumn{4}{c|}{\multirow{2}{*}{\begin{tabular}[c]{@{}c@{}}Stage=1\\ N=8 
  {      } N=12 {      }    N=16 {      }    N=20\end{tabular}}} &
  \multirow{2}{*}{\begin{tabular}[c]{@{}c@{}}Stage=2\\ N=32\end{tabular}} &
  \multirow{2}{*}{\begin{tabular}[c]{@{}c@{}}Stage=3\\ N=32\end{tabular}} \\
     & \multicolumn{4}{c|}{}          &       &       \\ \midrule \midrule
PSNR & 31.64 & 31.92 & 32.24 & 32.32 & 32.85 & 33.06 \\ \midrule
SSIM & 0.947 & 0.949 & 0.955 & 0.955 & 0.958 & 0.961 \\ \bottomrule
\end{tabular}
\label{tab:table5}
\end{table}

\subsection{Comparisons with residual dense block} 

Our RDAFB is designed based on hierarchical attentions, while RDB is designed based on hierarchical features. In this experiment, we compare our RDAFB with RDB \cite{Zhang2020} using one-stage RDAFNet as a backbone network. For a fair comparison, all models have the same number of convolution layers, and each model has a similar model size. The comparison experiments are performed on the GoPro dataset \cite{Nah2017}. The experiment results are shown in Table \ref{tab:table4}. When all models have 32 convolution layers, it can be seen that our RDAFB brings a favorable gain of 0.62dB PSNR over the RDB. We also compare the performance of RDB and RDAFB in cases of 64 convolution layers stacked, and RDAFB is still performs better than RDB. This experiment further indicates that the effective use of hierarchical attentions could make the deblurring network achieve excellent deblurrring performance.

\subsection{Why use multistage?} 

Using a large number of layers and larger filters, which help to increase the CNN’s receptive field and generalization ability, can improve the performance of computer vision or image processing tasks. However, these technologies create a suboptimal design for deblurring, since network performance does not always improve with the increasing network depth, and the effective receptive field of CNN is much smaller than the theoretical value \cite{Nah2017}. To understand this point, we show the performance of various numbers of RDAFBs stacked in the one-stage RDAFNet, two-stage RDAFNet and three-stage RDAFNet. As we can see in Table \ref{tab:table5}, although the performance improves with the number of RDAFBs in one-stage RDAFNet, the improvement becomes limited after 16 RDAFBs stacked. However, after we stacked RDAFBs in two-stage RDAFNet, the performance of the proposed model was significantly improved, as shown in Table \ref{tab:table5}. The three-stage network RDAFNet with the same numbers of RDAFBs (N=32) achieves a performance gain of 0.21dB PSNR compared with two-stage RDAFNet. 

\section{Conclusion}
In this paper, we introduce a new continuous cross-layer attention transmission (CCLAT) mechanism that makes full use of all hierarchical attentions through locally dense connections of attention maps. Based on the CCLAT mechanism, we designed a simple residual dense attention fusion block (RDAFB). In the RDAFB, the attention maps of preceding layers are densely connected with the current attention map, and then the connected attention maps are combined by the convolutional layers to obtain the fused current attention map. We also design an effective network (RDAFNet) for image deblurring, where the RDAFB serves as the basic building block. The experiments show that our RDAFNet model outperforms existing state-of-the-art methods.

\noindent
~\\
\textbf{Declaration of interests}
~\\

The authors declare that they have no conflict of interest.

\noindent
\\
\textbf{Acknowledgements}
~\\

This work is supported by the National Natural Science Foundation of China under Grant 61801337 and Grant 62171329




\bibliographystyle{elsarticle-num}
\bibliography{reference12.26}
\end{document}